\documentclass[conference]{IEEEtran}
\usepackage[total={6in, 8in}]{geometry}
\usepackage{hyperref}
\hypersetup{
    colorlinks=true,
    linkcolor=cyan,
    citecolor=cyan,
    filecolor=cyan,      
    urlcolor=cyan,
    pdftitle={Morphological Wobbling Can Help Robots Learn},
    pdfpagemode=FullScreen,
}

\usepackage{natbib}
\usepackage{amsmath,amssymb,amsfonts}
\usepackage{algorithmic}
\usepackage{graphicx}
\usepackage{textcomp}
\usepackage{xcolor}
\begin{document}

\title{Morphological Wobbling Can Help Robots Learn}

\author{\IEEEauthorblockN{Fabien C. Y. Benureau}
\IEEEauthorblockA{\textit{Okinawa Institute of Science and Technology}\\
Okinawa, Japan \\
0000-0003-4083-4512}
\and
\IEEEauthorblockN{Jun Tani}
\IEEEauthorblockA{\textit{Okinawa Institute of Science and Technology}\\
Okinawa, Japan \\
0000-0002-9131-9206}
}

\maketitle

\begin{abstract}
We propose to make the physical characteristics of a robot oscillate while it learns to improve its behavioral performance. We consider quantities such as mass, actuator strength, and size that are usually fixed in a robot, and show that when those quantities oscillate at the beginning of the learning process on a simulated 2D soft robot, the performance on a locomotion task can be significantly improved. We investigate the dynamics of the phenomenon and conclude that in our case, surprisingly, a high-frequency oscillation with a large amplitude for a large portion of the learning duration leads to the highest performance benefits. Furthermore, we show that \emph{morphological wobbling} significantly increases exploration of the search space.
\end{abstract}


\section{Introduction}

Most robots used today have a fixed morphology. The length of their limbs, the strength of their actuators, the mass of their parts is decided during their design and fixed thereafter. In that respect, most robots are similar to adult humans. And since adult humans can acquire skills, can \emph{learn}, it appears reasonable to expect that a fixed-morphology robot, given a sophisticated enough programming, should be perfectly suited to learn as well. Yet, it's difficult to fail to notice that humans acquire the most skills at a time when their morphology is \emph{not} fixed: during their childhood. It is tempting to argue that skill acquisition is at its peak during childhood because babies begin with many things to learn, and furthermore that morphological development happening at this time is a coincidence driven by physiological necessities, and in fact probably impedes the pace of skill acquisition. An other---opposing---theory is that morphological development crucially guides and helps skill acquisition in humans and animals. This is the idea explored by \emph{developmental robotics} \citep{Lungarella2003, Cangelosi2015}. This study explores this central question: can a  robot whose morphology is not fixed see their learning performance improved over a fixed-morphology one?

Let's start by dispelling any ambition at biological plausibility here. We are not going to make a robot's morphology grow from a baby body into an adult one. We explored this idea---in the context of evolutionary robotics---in a previous study \citep{Benureau2022}. Other studies have explored this as well \citep{NayaVarela2020,NayaVarela2020b}: a robot's learning performance can benefit from growing up. Designing a robot that grows, however, is far from trivial. Besides the many technical difficulties this represents, even choosing how it should grow is a though problem, especially as knowledge on that issue is scarce. In terms of efforts too, it may necessitate essentially designing several different robots for each of the stages (e.g. infant/kid/adult), and ways for the robot to transition from one to the other if development is gradual. All this conspires to make morphological development too hard and costly in many cases to actually use it in robots. Here, we explore a simpler alternative. Rather than growing our robots like animals, we \emph{wobble} the morphology of a robot around its adult shape; we apply, with each passing learning epoch, a sinusoidal perturbation of the morphological values of a robot such as its size, its mass, or actuator strength.

The idea of perturbating the target task or environment is not new and one example is Jakobi's minimal simulation approach \citep{Jakobi1997, Jakobi1998b}. The method advocates making any non-task-related part of a simulation \emph{noisy}, with the objective to make the discovered behaviors robust enough to experimental dimensions not crucial to the task so the robot can hopefully bridge the reality gap \citep{Koos2013}. In comparison to this approach, 
morphological wobbling target dimensions a priori deeply entangled with the task, such as the robot size or motor strength; it modifies the morphology \emph{away} from the target one. It seems, at first sight, an obstacle rather than a learning help.

Growing robots or morphologically developing robots  \citep{NayaVarela2021} is a subject that is still relatively niche, with the majority of the studies focused on the indeterminate growth of plant-like robots \citep{DelDottore2018, Corucci2017}.

A study \citep{Lungarella2002b} has specifically looked into how a nonlinear perturbation of a 12 DOFs robot affects performance when present or absent. They concluded, however, that its presence led to lower performance. This is also the conclusion of \citep{Bongard2011} when adding morphological development to a locomotion acquisition task. The solution for that study was to add morphological development during behavior, a technique that is used in a number of other studies \citep{Kriegman2017,Kriegman2018,Kriegman2018b,Bongard2011,Corucci2016}. Contrary to those studies, morphological wobbling does not change the morphology during a trial or learning epoch.

In recent studies, \citep{NayaVarela2020} and \citep{NayaVarela2020b} looked into how growth and gradual increase in range of motion affected learning in quadruped, hexapods, and octopods. In \citep{Benureau2022} we looked into how growth, increase in mass, or muscle strength (or all three combined) can affect learning in 2D  and 3D soft robots (the 2D one is same as this study). In all three studies, the developmental trajectory starts with a baby morphology that slowly grows, as learning progresses towards the adult morphology. All studies were able to show instances of development outperforming non-development. Rather than starting with a baby morphology that one has to design, morphological wobbling starts with the adult morphology and oscillates its morphological values.

\section{Method}

\subsection{Robots}

\begin{figure}[htbp]
\centerline{\includegraphics[width=0.5\textwidth]{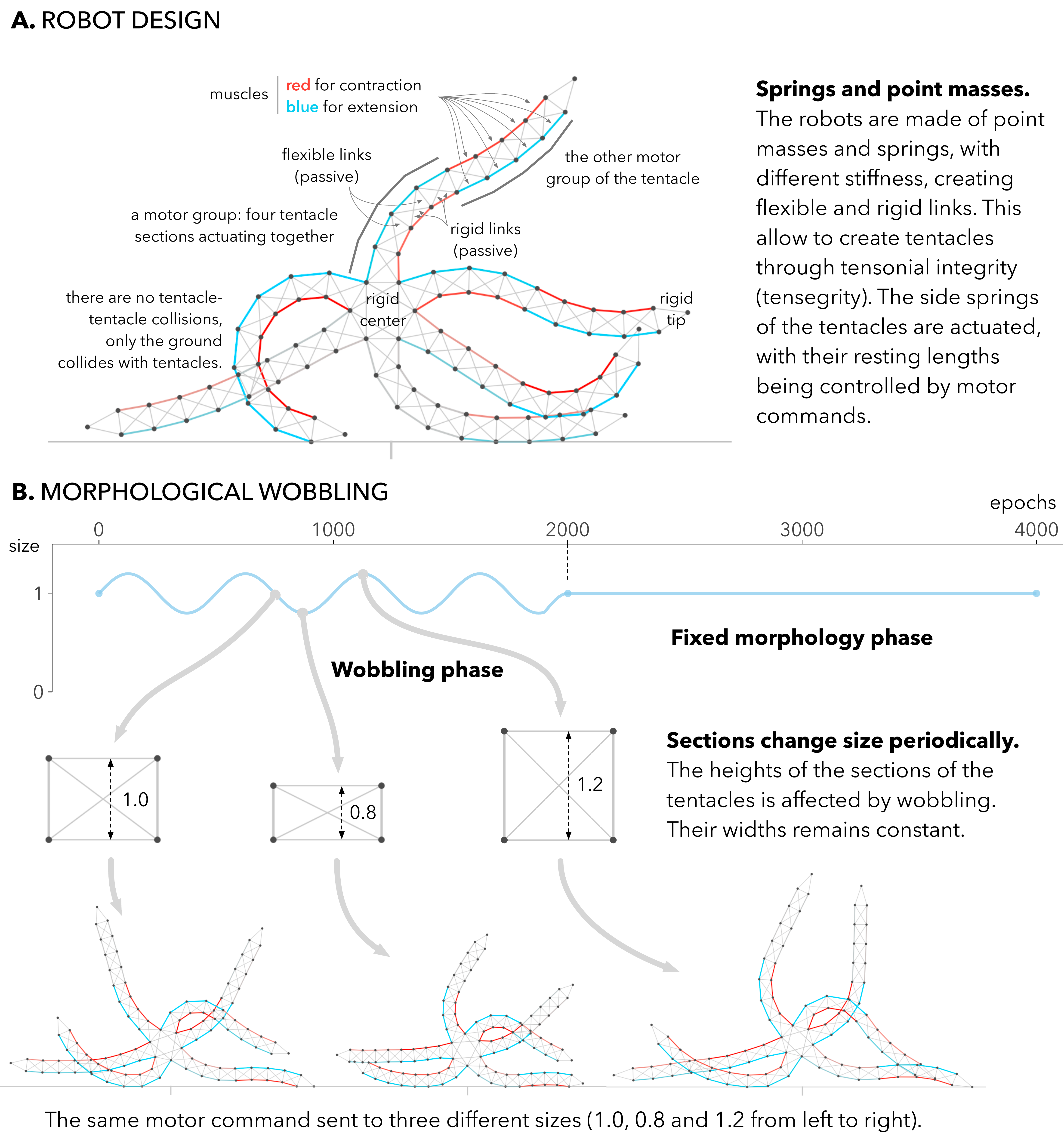}}
\caption{Morphological wobbling applies a sinus function to chosen morphological characteristics of the robot for the first half of training.}
\label{fig:robot}
\end{figure}

We consider a simulated 2D soft "starfish" robot, the same as the one in \citep{Benureau2022}. The robots are composed of six tentacles attached to a central body. Every part of the robot is composed of point masses linked together by springs. The springs can adopt a wide range of rigidity, enabling them to simulate rigid and flexible links alike. Some links allow their resting length to be modified by motor commands: those are the muscle of the robot. See Fig.~\ref{fig:robot}.A.

The robot has eight sections per tentacle. Each section possesses two actuated springs that act in an antagonistic manner: when one contracts, the other extends. Sections form two motor groups of four sections per tentacle. Inside a group, all sections receive the same motor commands: they contract and expand synchronously. All motors actuate according to a sinusoid signal. To be fully parameterized, they need a temporal offset and an amplitude. Using two scalars per motor group, and with 12 motor groups, a vector of 24 values suffices to fully parameterize the behavior of the robot.

\subsection{Trial \& Error Learning}

The task of our robot is to learn how to move. The performance is the number of body lengths (of the reference adult size, i.e. fixed, independent of wobbling) the robot is able to do over 60 seconds. We use a simple learning algorithm: trial and error. At each epoch, the robot tries 20 different behaviors. The five best are kept, and random perturbations of them are computed to generate 15 new behaviors that will, with the five kept behaviors, form the 20 behaviors of the next epoch. The robot is trained in this manner for 4000 epochs.


\subsection{Morphological Wobbling}

Our morphological wobbling consists in changing the morphological characteristics around the adult morphology values (see Fig.~\ref{fig:robot}.B). The morphological characteristics we consider are the mass of the nodes of the robot, the stiffness of the muscle springs, and the size (the height) of the tentacle sections. Typically, we apply a sinusoid perturbation of one of those values during the first 1900 epochs, with a given amplitude and perturbation. From epoch 1900 to 2000, the amplitude, whatever its value, is linearly decreased to zero. From epoch 2000 to 4000, the robot's morphology is fixed at the adult morphology values. For simplicity, we normalize all values so that the adult morphology values of mass, muscle stiffness, and tentacle section size is 1.0.

It is important to note that the morphology of the robot does not change during a learning trial or even during a whole epoch. It remains fixed and the is same for all trials. The morphology changes in-between epochs, during the first 2000 epochs. Another point is that our wobbling trajectory is not a function of learning performance, and does not change, for instance, when a performance threshold is reached.

\section{Results}

\begin{figure}[tb]
\centerline{\includegraphics[width=0.5\textwidth]{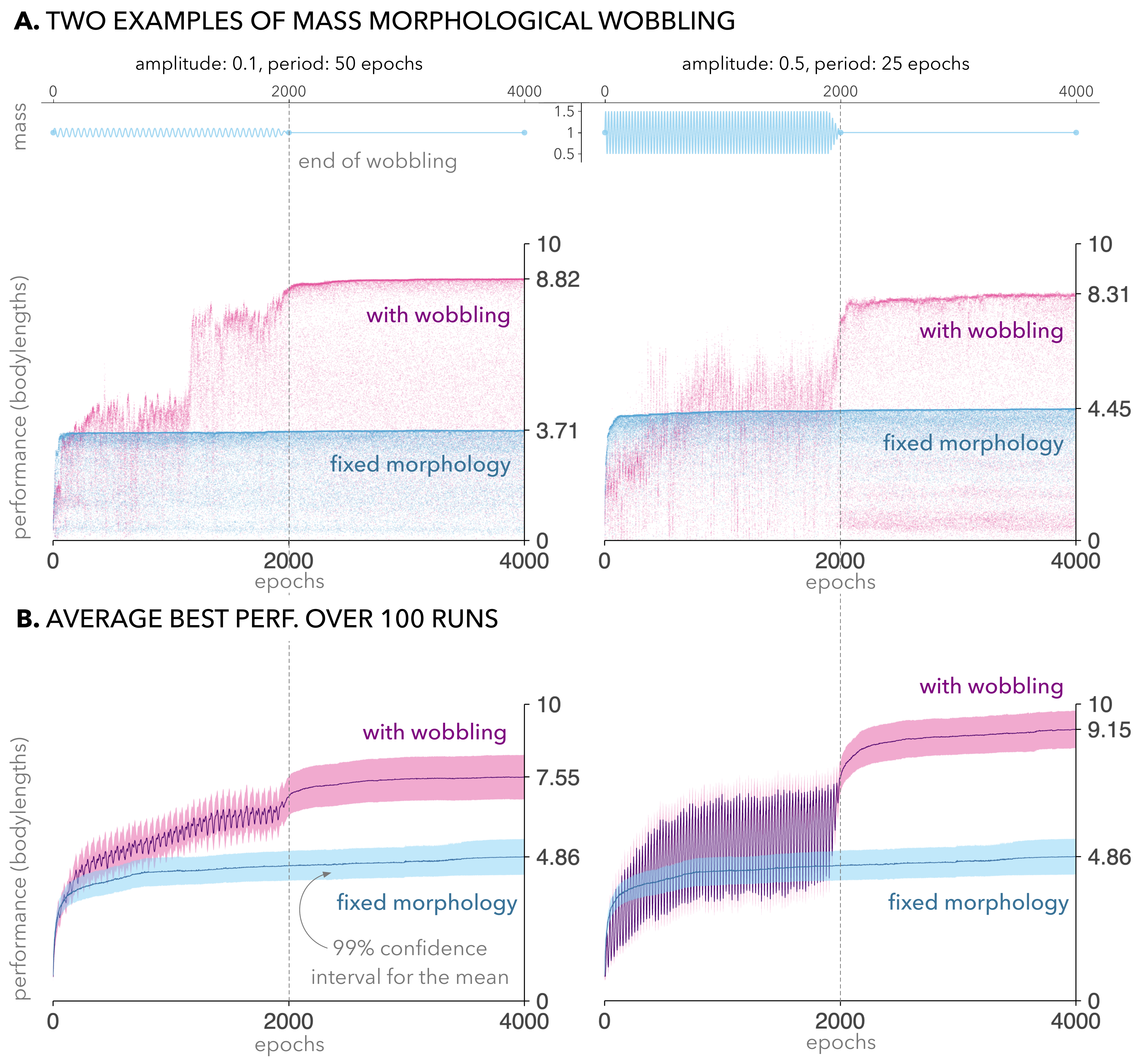}}
\caption{Morphological wobbling improves performance. \textbf{A.} Scatter plots of the trials of a run for the fixed-morphology and morphological wobbling conditions. Each dot is a trial. Futhermore, two oscillation conditions are shown and correspond to the left and right column, and extend to \textbf{B.}, which present the average of the best trial of each epoch, over 100 runs. The shaded area is the 99\% confidence interval.}
\label{fig:scatter}
\end{figure}

Fig.~\ref{fig:scatter} shows the impact of mass morphological wobbling for two conditions over the fixed-morphology condition. In Fig.~\ref{fig:scatter}.A, two selected runs are shown. We can observe that the performance during wobbling has a high variance that correlates with the wobbling oscillation. That is not surprising. Given similar behaviors, increasing or decreasing the mass of a robot will usually increase or decrease his locomotion performance. During wobbling, the performance oscillates at levels of performance inferior, similar, or superior to the fixed-morphology performance, with sudden changes to the mean. After wobbling ends, however, the performance of the wobbled robot settles to a value much higher than the fixed morphology.

Those results are validated when we look at the average performance of the best trial of each epoch across 100 runs in Fig.~\ref{fig:scatter}.B. We can see that the performance indeed oscillates precisely with the wobbling (else, it would get smoothed out by the average). And the final performance is significantly higher than the fixed morphology one, up to twice as much for amplitude 0.5 and a period of 25 epochs.

Those last parameters represent quite a high amplitude and a fast period. It is perhaps the most unexpected result of this study: we assumed that a moderate period and amplitude would reap the benefit of wobbling while giving the time to the learning algorithm to adapt. Yet---and this is the point of the next two sections---performance is generally increased the higher the amplitude and the shorter the period.

\subsection{Period and Wobbling Pauses}

\begin{figure}[htbp]
\centerline{\includegraphics[width=0.5\textwidth]{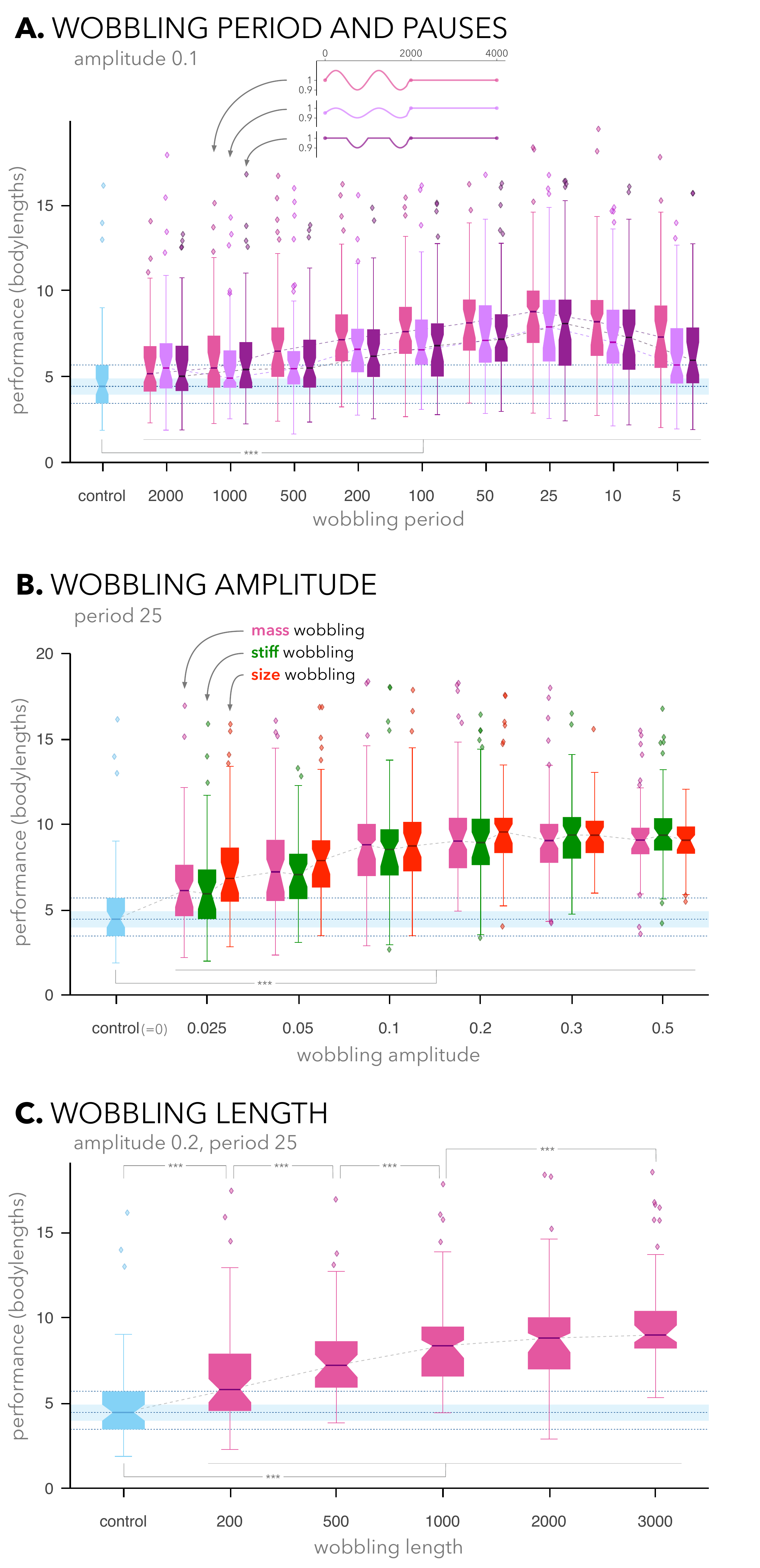}}
\caption{Robots should wobble for a long time (C.), at high frequency (A.) and high amplitude, regardless of the morphological dimension of wobbling (B.). Each boxplot is computed from 100 runs. The same set of 100 random seeds is used for each boxplot, creating paired experiments with the same initial populations. Boxplots display the first and third quantile, and the minimum and maximum value, or $1.5 \times \textrm{IQR}$, whichever is closer to the median; outliers are represented by diamonds, while notches show the 99\% confidence interval of the median. Significance is computed using a Wilcoxon signed-rank test after Shapiro-Wilk testing revealed that the differences between some of the paired fitnesses deviate significantly from normality. The significance threshold is set at 0.01. These conventions will be used for all boxplots of the article.}
\label{fig:many}
\end{figure}

To apply morphological wobbling, we must choose a few variable for the oscillations: the period, the amplitude, and how long they last. Let's start with the period, i.e., how fast the morphology wobbles. As stated above, we could expect that a slow wobbling, perhaps even interlaced with some fixed-morphology "rest" periods, would be best for learning. The results of Fig.~\ref{fig:many}.A show it is not the case for our robot. We consider three conditions across different period values. The first condition is the regular wobbling around the 1.0 mass value, with an amplitude of 0.1. The second is the same wobbling, with an upper bound at a value of 1.0; this creates a fixed value at 1.0 for half of the period. Finally, the third condition wobbles around a mean of 0.9, with an amplitude of 0.95 (the fixed value during the second half of learning is still 1.0). This has the same range as condition 2 without any fixed-morphology portions. What we observe is that the period offering the most performance is 25 epochs, rather than the slower period we expected. A shorter period than 25 does lead to some modest loss of performance, suggesting that the relative proximity of successive morphologies along the oscillation range is of some importance. Stopping to wobble periodically as in condition 2 does not offer any advantage over condition 3 and has a consistently inferior mean performance over condition 1. Comparing condition 1 and 3 further suggests that having the mean of the oscillation be the value of the target morphology might be important, but this conclusion is muddled by the difference of amplitude, especially in light for the results of the next section.

\subsection{Amplitude and Morphological Characteristics}

The other crucial parameter of wobbling it its amplitude, i.e.: how much to wobble? Fig.~\ref{fig:many}.B shows both the effect of amplitude and of wobbling different morphological characteristics on performance for a period of 25. The performance increases as the amplitude of wobbling increases, until 0.3. After, further increases in amplitude do not increase performance significantly. This is again unexpected. The higher the amplitude, the further away the robot is from the target, adult morphology. Behaviors that are effective on a morphology are rarely effective on a highly different one, with robotic search space often dense with non-linearity and discontinuities. This could easily lead to a situation where a learning algorithm cannot keep up with the fast-oscillating environmental changes caused by changing morphology, and regularly discard good solutions because half of the learning time is spent on morphology significantly different from the target one. Yet, we do not observe this here: a high amplitude of oscillation is beneficial, and though we observe diminishing returns, we do not observe any adverse effects of high amplitudes.

Interestingly, even for qualitatively different forms of morphological wobbling---node mass, muscle stiffness, or size---, the performance is impacted the same. This non-specificity of the wobbling is surprising. We would expect that some morphological characteristics among the ones we considered would be coupled more tightly to the learning performance than others.

\subsection{Wobbling Length}

A final major question is: how long should the wobbling lasts? Fig.~\ref{fig:many}.C tells us that even a short wobbling phase of 200 epochs brings a significant learning performance benefit. And the longer the wobbling phase is, the higher the performance benefit. This is not a trivial result. Where is that dynamic coming from? To answer this, we need to start understanding why morphological wobbling works.

\subsection{Morphological Wobbling Fosters Exploration}

Given a task, a robot learns by exploring a behavioral space and finding behaviors that solve the task as best as possible. One danger of learning is being trapped in a local extremum of the behavioral space and spending time improving a specific solution when better ones exist in other areas of the space. Such premature convergence is an inevitable risk of any learning happening in a behavioral space that cannot be exhaustively explored. Many algorithmic techniques exist in machine learning to minimize that risk.

Morphological wobbling is an \emph{embodied} technique to reduce that risk. Rather than finding clever ways to explore a behavioral space, morphological wobbling \emph{modifies the behavioral space}. It is difficult to stay trapped in a valley of the behavioral space when the valley disappears under you.

One way to see the relationship between a learning algorithm and a dynamic environment is to see the learning algorithm as having to constantly play catch up with the environmental changes. Another is to consider that the environment pushes the learning process around, especially out of stable attractors.

\begin{figure*}[tb]
\centerline{\includegraphics[width=0.9\textwidth]{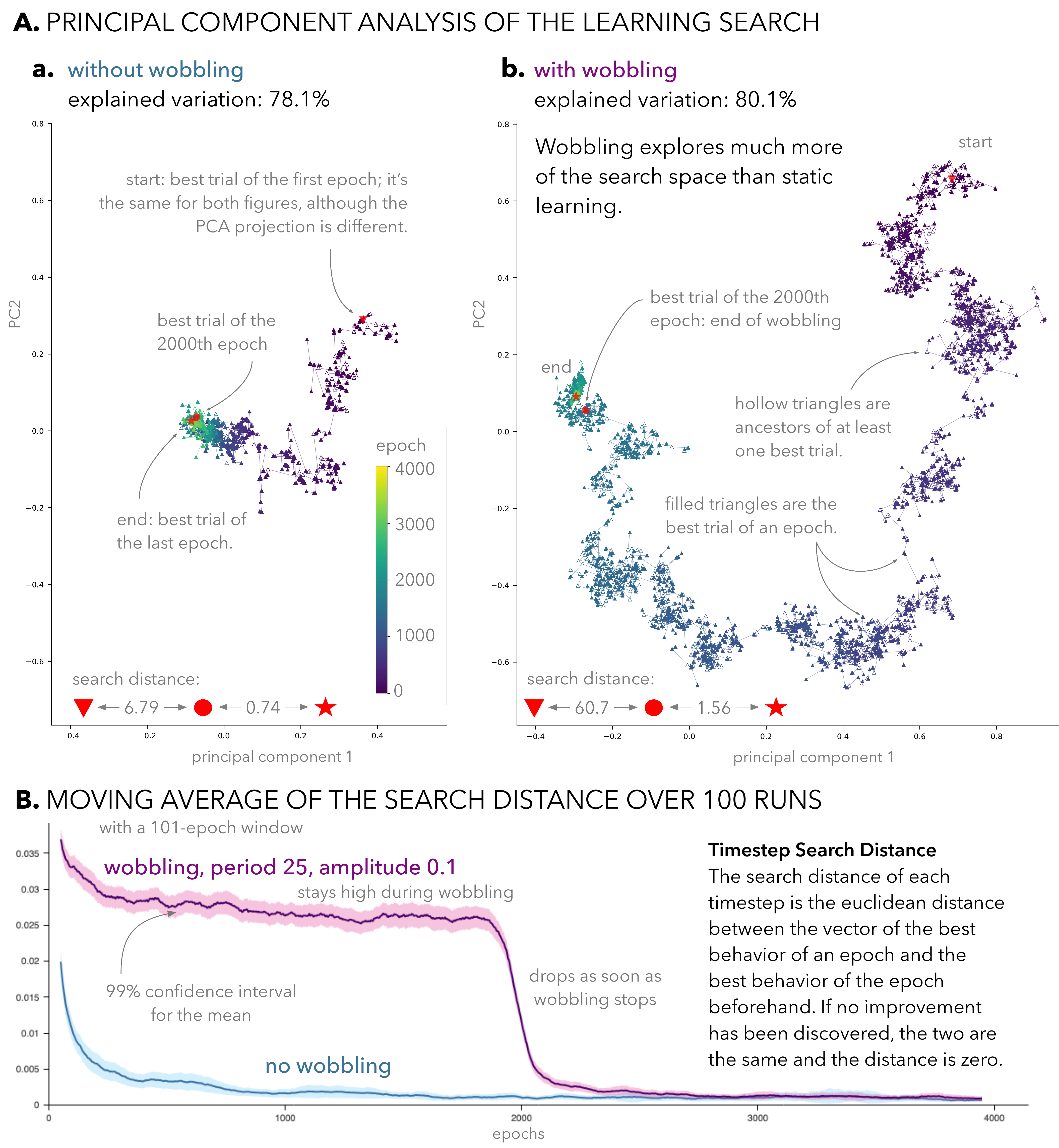}}
\caption{Wobbling makes the learning process explore the search space much more. \textbf{A.} Starting from the same initial behavior, we perform a 2D principal component analysis of the search through the behavioral space. Each behavior except the initial one is derived from another and we can therefore draw a graph of the behavior. Here we represent only the best behavior of each epoch and its ancestors. The starting point is a large inverted red triangle, the halfway point (epoch 2000) is a red disk, and the endpoint is a red star. The search distance, as the sum of the euclidean distance between each consecutive behavior (encoded as a 24-scalar vector), is given between those points for the fixed-morphology and morphological wobbling conditions. The two PCAs are made independently for each condition (the projection axes are different). \textbf{B.} We confirm that the search distance dynamics are shared over the 100 runs. We show the moving average of the \emph{timestep search distance} (the euclidean distance from the best behavior genotype vector from one epoch to the next) over 100 runs and a 101-epoch window.} 
\label{fig:pca}
\end{figure*}

Fig.~\ref{fig:pca} gives us an indication that this is indeed what is happening here. We started a robot with a single behavior. The robot learns to modify this behavior to improve its locomotion. In one condition, the robot has a fixed morphology, and in the other, the robot undergoes morphological wobbling. Since at each epoch the robot tries new behavior by modifying the existing ones, we can measure the number of modifications that we explored from the initial behavior to the best behavior of the last epoch. We obtain a \emph{search distance} as the sum of the euclidean norm of all these modifications. In the run of Fig.~\ref{fig:pca}, the fixed morphology covered a distance of 6.79  over the first 2000 epochs (6.66±0.84 (99\% CI) over 100 runs); the wobbling robot covered a distance of 60.7 (58.23±1.49 over 100 runs) in the same timeframe. As soon as the wobbling stops, the search distance drops sharply, to cover a distance of 0.74 (1.81±0.45) for the fixed morphology over the next 2000 epochs and 1.56 (3.33±0.65) for the wobbling one.

Another point to gather from Fig.~\ref{fig:pca} is that the interaction between a morphology that dislodges learning from the local extremum and learning that gravitates toward better behaviors does not seem to lead to a stationary dynamic in the search space. The search does not seem to visit again and again the same place of the search space, as could be expected of a repetitive, sinusoidal, perturbation: the search \emph{moves around} in the behavioral space, as shown through the PCA. This remains an informal observation that needs further evidence to be confirmed.

\section{Discussion}

There are many limitations to the current study. We have only one type of robot, the robots are in 2D, in simulation, the task is locomotion on a perfectly flat surface, there are no sensors, no neural networks. They all warrant further study and expanding the work to figure out how much, if at all, this result generalizes to different robots, tasks, and environments. 

There's one aspect that warrants an extended discussion: the fixed morphology converges \emph{quickly}, as is evidenced in Fig.~\ref{fig:scatter}.A. First, this fast convergence it probably a reason why a short period and therefore fast oscillations produce good results in this study. It could lead to the conclusion that morphological wobbling is effective here because it prevents the premature convergence of a crude trial-and-error learning algorithm; in contrast, more sophisticated learning algorithms are not susceptible to such a premature convergence and, therefore, would not get any benefit from morphological wobbling.

It's a valid point. Except that premature convergence can befall even sophisticated learning algorithms, and that the additional help of morphological wobbling, although not useful in many situations and tasks, might be crucial in some others. But a better justification here is to argue that this quick convergence of the task may actually capture some of the dynamics at play in the biological world.

Indeed, there is evidence that a simple trial and error algorithm may suffice to acquire new sensorimotor skills in biology, simply because the search space has a lot of good solutions, and we may be able to converge on one of them quickly, wherever our starting point is \citep{Raphael2010, Loeb2012}. Quickly finding a good-enough solution to a problem is useful for survival in the animal kingdom. Yet it could prevent finding a better solution that might represent a negative selection pressure over time, especially if other members of the population discover it. Here, morphological change, i.e. physical growth, may help improve over time a behavior that converged quickly. We would have both advantages: a fast convergence toward a useful behavior and a high performing behavior by the time the animal reaches adulthood.

One problematic aspect of robotic morphological development is the difficulty in implementing it in practice. Designing a robot that grows is complicated and run contrary to many of the other engineering constraints of robot design. Even in simulation, designing a developmental path is far from trivial, and may present a difficulty equal or superior to the design of the robot itself. Morphological wobbling removes a lot of those difficulties. Many of our experiments have dealt with changing the mass of our robots. In simulation, this can be achieved directly. Even more simply, a similar effect can be achieved by wobbling the gravity of the simulation, which amounts to changing the value of a single vector, and removes the need to recompute inertia matrices. With real robots, changing gravity is not possible and modifying the mass---or the size---can be challenging or impossible. A simple intervention, however, is to change the motor strength by modifying the max torque for instance or by using specialized actuators \citep{Vu_2013}. In all those cases, the amplitude provides an easy way to constraint the wobbling within the physical limits of the robot. With only a handful of parameters to consider---period, amplitude, duration---morphological wobbling is a comparatively easy implementation of morphological development.

\section*{Conclusion}

We presented morphological wobbling and showed that deliberately applying a sinusoidal perturbation to the morphology of a robot can increase its learning performance. We showed that, surprisingly, in our case study, a high frequency and high amplitude perturbation applied for a high number of epochs provided the best results. We analyzed how exploration was affected by morphological wobbling, showing that the distance covered in the search space increased significantly during the oscillations.

A lot of work remains to investigate how successfully morphological wobbling can be used in different learning robots and across environments and tasks. Whether it proves to be widely applicable or only applicable to a handful of cases, such as our 2D robots, even a failure is interesting as it touches on the relationship between development and learning. If morphological wobbling improves the performance of some robots but not others, understanding why may yield precious insights into how development affects us.

\section*{Acknowledgment}

We are grateful for the help and support provided by the Scientific Computing and Data Analysis section of Research Support Division at OIST.

\section*{Source code}

The source code and full simulation logs to reproduce the results, regenerate the plots, and recompute the numerical figures (distances, confidence interval, significance, explained variation of PCA) is available at: \url{https://doi.org/10.5281/zenodo.6513360}.

\bibliography{references}

\end{document}